\documentclass[10pt,twocolumn,letterpaper]{article}
\usepackage[pagenumbers]{wacv} 

\usepackage{latexsym}
\usepackage{amssymb}
\usepackage{amsmath}
\usepackage{amsthm}
\usepackage{booktabs}
\usepackage{enumitem}
\usepackage{graphicx}
\usepackage{color}
\usepackage[ruled, vlined]{algorithm2e}
\usepackage{multirow}
\usepackage{threeparttable}
\usepackage{graphicx}
\usepackage{amsmath}
\usepackage{amssymb}
\usepackage{booktabs}
\usepackage{morefloats}
\extrafloats{100}
\usepackage{etex}
\usepackage[pagebackref,breaklinks,colorlinks]{hyperref}

\usepackage[capitalize]{cleveref}
\crefname{section}{Sec.}{Secs.}
\Crefname{section}{Section}{Sections}
\Crefname{table}{Table}{Tables}
\crefname{table}{Tab.}{Tabs.}

\def\wacvPaperID{2330} 
\def\confName{WACV}
\def\confYear{2025}

\begin{document}

\title{Hierarchical Adaptive Expert for Multimodal Sentiment Analysis}

\author{Jiahao Qin\\
Xi'an Jiaotong-Liverpool University\\
Suzhou, China\\
{\tt\small Jiahao.qin19@gmail.com}
\and
Feng Liu\\
Shanghai Jiao Tong University\\
Shanghai, China\\
{\tt\small lsttoy@163.com}
\and
Lu Zong\\
Xi'an Jiaotong-Liverpool University\\
Suzhou, China\\
{\tt\small lu.zong@xjtlu.edu.cn}
}
\maketitle

\begin{abstract}
  Multimodal sentiment analysis has emerged as a critical tool for understanding human emotions across diverse communication channels. While existing methods have made significant strides, they often struggle to effectively differentiate and integrate modality-shared and modality-specific information, limiting the performance of multimodal learning. To address this challenge, we propose the Hierarchical Adaptive Expert for Multimodal Sentiment Analysis (HAEMSA), a novel framework that synergistically combines evolutionary optimization, cross-modal knowledge transfer, and multi-task learning. HAEMSA employs a hierarchical structure of adaptive experts to capture both global and local modality representations, enabling more nuanced sentiment analysis. Our approach leverages evolutionary algorithms to dynamically optimize network architectures and modality combinations, adapting to both partial and full modality scenarios. Extensive experiments demonstrate HAEMSA's superior performance across multiple benchmark datasets. On CMU-MOSEI, HAEMSA achieves a 2.6\% increase in 7-class accuracy and a 0.059 decrease in MAE compared to the previous best method. For CMU-MOSI, we observe a 6.3\% improvement in 7-class accuracy and a 0.058 reduction in MAE. On IEMOCAP, HAEMSA outperforms the state-of-the-art by 2.84\% in weighted-F1 score for emotion recognition. These results underscore HAEMSA's effectiveness in capturing complex multimodal interactions and generalizing across different emotional contexts. The code will be available on GitHub soon.
\end{abstract}

\section{Introduction}
Multimodal sentiment analysis has emerged as an important research area in recent years, aiming to interpret human emotions and sentiments by integrating information from multiple modalities, such as text, audio, and visual cues \cite{cambria_review_2017,kaur_multimodal_2019}. This field has significant implications for various applications, including human-computer interaction, customer analysis, and social media computing \cite{poria_review_2017,zadeh_multimodal_2020}.

\begin{figure}[!ht]
\centering
\includegraphics[scale=1.0]{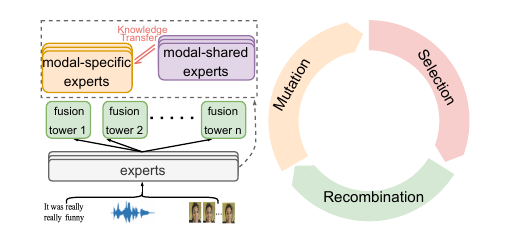}
\caption{Overview of the Hierarchical Adaptive Expert Network (HAEN) and the evolutionary optimization process. The left portion illustrates the hierarchical structure, where modal-specific experts receive knowledge transfers from modal-shared experts to enable cross-modal knowledge fusion and multi-task optimization. The right portion depicts the evolutionary algorithm’s key steps—Selection, Recombination, and Mutation—adapting network configurations over iterative generations.}
\label{fig:small}
\end{figure}

Recent advances in deep learning have enabled the development of more sophisticated multimodal sentiment analysis models \cite{yu_learning_2021,zadeh_tensor_2017}. These models attempt to capture complex interactions among modalities and leverage complementary information to improve sentiment prediction accuracy \cite{ghosal_contextual_2018,mai_analyzing_2019,app14177720}. However, existing approaches often struggle to effectively combine modality-specific representations and fully capture the intricate dynamics of multimodal data \cite{hazarika_misa_2020,sun_progressively_2022,qin2024zoomshiftneed,qin2024stepfusionlocalglobal,qinicic2024}. To address these challenges, researchers have explored various approaches, including hierarchical architectures for learning representations at different granularities \cite{majumder_hierarchical_2019,gao2018neural}, and multi-task learning (MTL) techniques for leveraging knowledge from related tasks \cite{akhtar_multi-task_2019, majumder_sentiment_2019}.

However, designing effective hierarchical architectures and MTL frameworks often requires significant manual effort and domain expertise. This is where evolutionary algorithms may offer a potential solution. These algorithms, inspired by biological evolution, have been successfully applied to various aspects of deep learning, such as architecture search and hyperparameter optimization.

In the context of multimodal sentiment analysis, evolutionary algorithms present several potential advantages: They can automate the search for optimal architectures and modality combinations, potentially reducing the need for extensive manual design; Their population-based search and recombination mechanisms allow for the exploration of diverse architectural variants; They offer the potential to balance multiple optimization objectives, such as prediction accuracy, modality fusion effectiveness, and computational efficiency.

Inspired by these insights, we propose the Hierarchical Adaptive Expert for Multimodal Sentiment Analysis (HAEMSA) framework. HAEMSA leverages evolutionary algorithms to develop a hierarchical structure of specialized subnetworks, each serving as an adaptive expert. These experts collaboratively learn to represent and combine information from multiple modalities at varying levels of granularity, while also informing the learning process of modality-specific components. Figure \ref{fig:small} demonstrates the impact of different population sizes in the evolutionary process on model performance, highlighting the importance of this approach.
The main contributions of our work are as follows:
\begin{itemize}
\item We introduce HAEMSA, a novel framework that uses evolutionary algorithms to optimize hierarchical architectures for multimodal sentiment analysis. This approach enables the capture of intricate inter-modal interactions and dependencies at multiple granularities, improving upon existing flat or manually designed architectures.
\item We develop a targeted knowledge transfer mechanism that facilitates the sharing of learned representations across modalities and related tasks. This mechanism enhances multimodal learning by leveraging shared patterns and relationships, leading to more robust and generalizable representations.
\item We integrate HAEMSA with a multi-task learning approach, enabling the model to simultaneously learn from multiple sentiment analysis tasks. This integration allows for the capture of task interdependencies, resulting in improved performance across various sentiment analysis subtasks.
\item We conduct comprehensive experiments on three benchmark datasets: CMU-MOSI, CMU-MOSEI \cite{zadeh_multi-attention_2018}, and IEMOCAP \cite{busso2008iemocap}. Our results demonstrate HAEMSA's effectiveness across various multimodal sentiment analysis tasks. Specifically, we achieve improvements of up to 6.3\% in 7-class accuracy on CMU-MOSI, 5.6\% on CMU-MOSEI compared to state-of-the-art methods, and a 2.84\% increase in weighted-F1 score for emotion recognition on IEMOCAP. These consistent improvements across diverse datasets underscore the robustness and generalizability of our approach.
\end{itemize}


\section{Related Work}
\subsection{Multimodal Sentiment Analysis}
Multimodal sentiment analysis has seen significant advancements with the advent of deep learning. Recent approaches focus on capturing complex interactions among modalities \cite{poria2017context,hazarika2018icon}. Notable works include the contextual inter-modal attention framework by Ghosal et al. \cite{ghosal_contextual_2018} and the Multi-attention Recurrent Network (MARN) by Zadeh et al. \cite{zadeh_multi-attention_2018}.

Despite these advancements, challenges persist in effectively combining modality-specific representations and generalizing to unseen data \cite{hazarika_misa_2020,sun2022learning}. To address these limitations, researchers have explored multi-task learning \cite{akhtar_multi-task_2019} and adversarial training \cite{han2021improving} to improve model generalization and adaptability.

\subsection{Evolutionary Approaches and Hierarchical Architectures}
Evolutionary algorithms have shown promise in multimodal learning, particularly in neural architecture search \cite{liu_hierarchical_2018,real_regularized_2019}. These methods automatically discover optimal architectures by evolving a population of candidates based on their performance. In multimodal sentiment analysis, evolutionary approaches offer the potential to reduce manual design efforts \cite{martinez_evolutionary_2022}. Hierarchical architectures have been explored to capture the structure of multimodal data at different granularities \cite{yang_hierarchical_2016,majumder_dialoguernn_2019}. These typically consist of modality-specific layers followed by cross-modal layers to capture inter-modal interactions \cite{liu_hierarchical_2021}.

\subsection{Multi-task Learning in Sentiment Analysis}
Multi-task learning (MTL) has been widely adopted in sentiment analysis to leverage knowledge from related tasks \cite{zhang_survey_2021}. In multimodal sentiment analysis, MTL approaches often incorporate hierarchical structures to capture cross-modal interactions effectively \cite{yu_multimodal_2021}. However, designing effective MTL architectures for multimodal sentiment analysis remains challenging. It requires balancing contributions from different modalities and tasks while capturing their intricate interactions \cite{sun2022learning}. Careful selection and weighting of auxiliary tasks are crucial to avoid negative transfer and ensure positive knowledge sharing \cite{ruder_overview_2017}.

Our proposed Hierarchical Adaptive Expert for Multimodal Sentiment Analysis (HAEMSA) framework addresses these challenges by integrating evolutionary algorithms with multi-task learning. HAEMSA evolves a hierarchical structure of specialized subnetworks, optimizing representations at various modality granularities. This approach enables effective learning of cross-modal interactions and task dependencies, advancing the state-of-the-art in multimodal sentiment analysis and emotion recognition.

\begin{figure*}[!ht]
\centering
\includegraphics[width=0.95\textwidth]{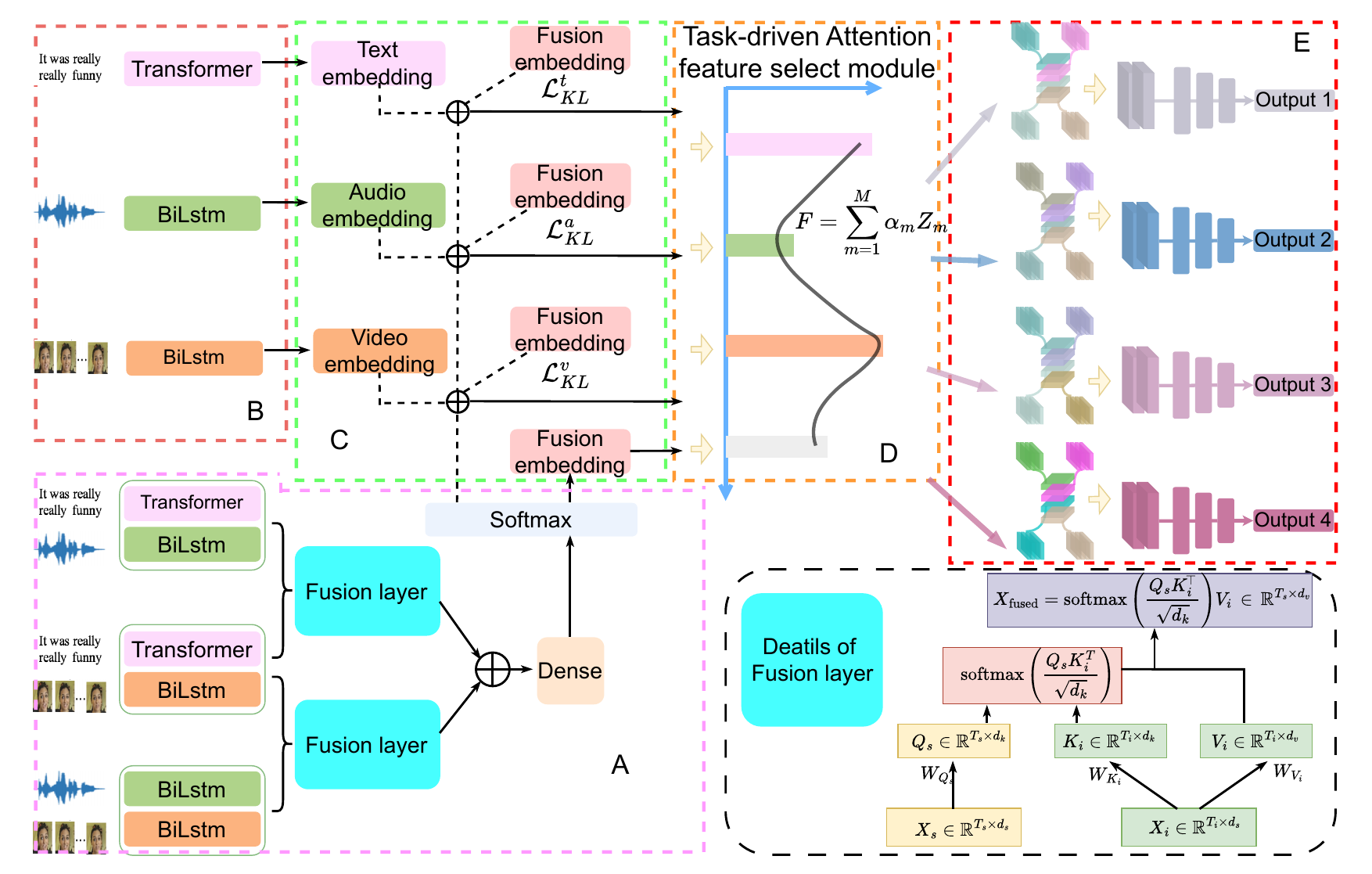}
\caption{Overview of the Hierarchical Adaptive Expert Network (HAEN) for MSA. The network consists of five main components: (A) an evolutionary adaptive modal-shared expert with a hierarchical structure that learns granular representations, (B) modality-specific experts that handle single-modality information, (C) collaborative cross-modal integration, (D) attention-based task-driven feature selection module, and (E) an adaptive tower network that utilizes task-driven gates for multi-task learning to capture sentiment analysis task dependencies.}
\label{fig:overview}
\end{figure*}


\section{Methodology}
In this section, we introduce the overall architecture and module details of the Hierarchical Adaptive Expert for Multimodal Sentiment Analysis (HAEMSA) framework for multi-modal sentiment analysis. Sequentially, we describe the hierarchical expert network structure, the evolution-based adaptive structural parameter optimization process, and the cross-modal, cross-task knowledge transfer mechanism.

\subsection{Hierarchical Adaptive Expert Network}
The hierarchical expert structure consists of multiple levels of experts that learn representations at different levels of granularity. Let $\mathbf{x}_t$, $\mathbf{x}_a$, and $\mathbf{x}_v$ denote the input features for the text, audio, and visual modalities, respectively. The modality-specific representations learned by the fine-grained experts can be expressed as:
\begin{equation}
\mathbf{h}_t = f_t(\mathbf{x}_t; \theta_t)
\end{equation}
\begin{equation}
\mathbf{h}_a = f_a(\mathbf{x}_a; \theta_a)
\end{equation}
\begin{equation}
\mathbf{h}_v = f_v(\mathbf{x}_v; \theta_v)
\end{equation}
where $f_t$, $f_a$, and $f_v$ are the fine-grained expert functions parameterized by $\theta_t$, $\theta_a$, and $\theta_v$, respectively, and $\mathbf{h}_t$, $\mathbf{h}_a$, and $\mathbf{h}_v$ are the resulting modality-specific representations.
The modality-shared representations learned by the unified experts can be expressed as:
\begin{equation}
\mathbf{h}_s = f_s([\mathbf{x}_t; \mathbf{x}_a; \mathbf{x}_v]; \theta_s)
\end{equation}
where $f_s$ is the unified expert function parameterized by $\theta_s$, $[\cdot;\cdot;\cdot]$ denotes concatenation, and $\mathbf{h}_s$ is the resulting modality-shared representation.
The hierarchical experts learn to fuse the modality-specific and modality-shared representations at different levels of granularity. Let $\mathbf{h}_l^{(i)}$ denote the fused representation at the $i$-th level of the hierarchy, where $l \in {t, a, v, s}$ indicates the modality. The fused representations can be computed as:
\begin{equation}
\mathbf{h}_l^{(i)} = g_l^{(i)}([\mathbf{h}_l^{(i-1)}; \mathbf{h}_s^{(i-1)}]; \phi_l^{(i)})
\end{equation}
where $g_l^{(i)}$ is the fusion function at the $i$-th level for modality $l$, parameterized by $\phi_l^{(i)}$, and $\mathbf{h}_l^{(i-1)}$ and $\mathbf{h}_s^{(i-1)}$ are the fused representations from the previous level.
\begin{table*}[h]
\centering
\caption{Performance comparison on CMU-MOSEI and CMU-MOSI datasets}
\label{tab:overall_performance}
\resizebox{\textwidth}{!}{%
\begin{tabular}{lcccccccc}
\hline
\multirow{2}{*}{Method} & \multicolumn{4}{c}{CMU-MOSEI} & \multicolumn{4}{c}{CMU-MOSI} \\
& Acc-7(\%) & Acc-5(\%) & Acc-2(\%) & MAE & Acc-7(\%) & Acc-5(\%) & Acc-2(\%) & MAE \\
\hline
TFN (2018) & 50.2 & - & 82.5 & 0.593 & 34.9 & - & 80.8 & 0.901 \\
LMF (2018) & 48.0 & - & 82.0 & 0.623 & 33.2 & - & 82.5 & 0.917 \\
MulT (2019) & 52.6 & 54.1 & 83.5 & 0.564 & 40.4 & 46.7 & 83.4 & 0.846 \\
Self-MM (2021) & 53.6 & 55.4 & 85.0 & 0.533 & 46.4 & 52.8 & 84.6 & 0.717 \\
MMIM (2021) & 53.2 & 55.0 & 85.0 & 0.536 & 46.9 & 53.0 & 85.3 & 0.712 \\
TFR-Net (2021) & 52.3 & 54.3 & 83.5 & 0.551 & 46.1 & 53.2 & 84.0 & 0.721 \\
AMML (2022) & 52.4 & - & 85.3 & 0.614 & 46.3 & - & 84.9 & 0.723 \\
EMT (2024) & 54.5 & 56.3 & 86.0 & 0.527 & 47.4 & 54.1 & 85.0 & 0.705 \\
CRNet (2024) & 53.8 & - & 86.2 & 0.541 & 47.4 & - & 86.4 & 0.712 \\
\hline
HAEMSA (Ours) & \textbf{57.1} & \textbf{58.2} & \textbf{86.7} & \textbf{0.482} & \textbf{53.7} & \textbf{55.2} & \textbf{86.9} & \textbf{0.654} \\
\hline
\end{tabular}%
}
\end{table*}

\subsection{Evolutionary Optimization for Automated MTL}
The HAEMSA framework incorporates multi-task learning to leverage the complementary information from related sentiment analysis tasks. Let $\mathcal{T}$ denote the set of tasks, which includes sentiment classification, emotion recognition, and sentiment intensity prediction. For each task $t \in \mathcal{T}$, a task-specific loss $\mathcal{L}_t$ is defined based on the ground-truth labels and the predicted outputs.
The multi-task learning objective is formulated as a weighted sum of the task-specific losses:
\begin{equation}
\mathcal{L}_\text{MTL} = \sum{t \in \mathcal{T}} w_t \mathcal{L}_t
\end{equation}
where $w_t$ is the weight assigned to task $t$.
The weights $w_t$ are learned adaptively during the training process using a task-specific attention mechanism. The attention mechanism learns to assign higher weights to the tasks that are more relevant for sentiment analysis, allowing the model to focus on the most informative tasks.
The total loss for the HAEMSA framework is a combination of the multi-task learning loss and the knowledge transfer loss:
\begin{equation}
\mathcal{L}_\text{total} = \mathcal{L}_\text{MTL} + \beta \mathcal{L}_\text{KT}
\end{equation}
where $\mathcal{L}_\text{KT}$ is the knowledge transfer loss, and $\beta$ is a hyperparameter that balances the contributions of the multi-task learning and knowledge transfer objectives.
By jointly optimizing the multi-task learning and knowledge transfer objectives, the HAEMSA framework effectively leverages the complementary information from related tasks and the knowledge captured by the unified experts, leading to improved sentiment analysis performance.

The structure and weights of the hierarchical experts are optimized through an evolutionary process. Let $\mathcal{P}$ denote the population of candidate architectures, where each individual $p \in \mathcal{P}$ represents a specific configuration of the hierarchical experts. The evolutionary process iteratively updates the population to discover optimal architectures and modality combinations.
The fitness of each individual $p$ is evaluated based on its performance on a validation set $\mathcal{D}\text{val}$. The fitness function $\mathcal{F}(p)$ can be expressed as:
\begin{equation}
\mathcal{F}(p) = \frac{1}{|\mathcal{D}\text{val}|} \sum_{(\mathbf{x}, y) \in \mathcal{D}_\text{val}} \mathcal{L}(p(\mathbf{x}), y)
\end{equation}
where $\mathcal{L}$ is a loss function that measures the discrepancy between the predicted sentiment $p(\mathbf{x})$ and the ground-truth sentiment $y$.
The evolutionary process employs genetic operators, such as mutation and crossover, to generate new individuals. The mutation operator introduces random perturbations to the architecture and weights of an individual, while the crossover operator combines the genetic information of two parent individuals to create offspring.
Let $p_i$ and $p_j$ be two parent individuals selected for crossover. The offspring $p_o$ is generated as follows:
\begin{equation}
p_o = \alpha \cdot p_i + (1 - \alpha) \cdot p_j
\end{equation}
where $\alpha \in [0, 1]$ is a random weight that determines the contribution of each parent.
The mutation operator applies random modifications to an individual $p$. The modified individual $p'$ is obtained as:
\begin{equation}
p' = p + \mathcal{N}(0, \sigma)
\end{equation}
where $\mathcal{N}(0, \sigma)$ is a Gaussian noise with mean 0 and standard deviation $\sigma$.
The selection of individuals for the next generation is based on their fitness values. We employ tournament selection, where $k$ individuals are randomly sampled from the population, and the one with the highest fitness is selected as a parent for the next generation.
The evolutionary process continues for a fixed number of generations or until a convergence criterion is met. The fittest individual in the final population represents the optimized architecture and weights of the hierarchical experts.

\subsection{Cross-Modal Joint Optimization}
The HAEMSA framework employs a joint optimization approach to effectively integrate knowledge across modalities and tasks. This approach combines knowledge transfer between unified and modality-specific experts with a multi-task learning objective.
\subsubsection{Knowledge Transfer Optimization}
The knowledge learned by the unified experts is transferred to the modality-specific experts to guide their learning. The knowledge transfer is achieved by minimizing the Kullback-Leibler (KL) divergence between the output probability distributions of the unified experts and the modality-specific experts.
The KL divergence loss for the text modality can be expressed as:
\begin{equation}
\mathcal{L}_\text{KL}^t = \sum_i p_s(y_i|\mathbf{x}_i) \log \frac{p_s(y_i|\mathbf{x}_i)}{p_t(y_i|\mathbf{x}_i)}
\end{equation}
where $p_s(y_i|\mathbf{x}_i)$ and $p_t(y_i|\mathbf{x}_i)$ are the output probability distributions of the unified expert and the text-specific expert, respectively, for the $i$-th sample $\mathbf{x}_i$.
Similarly, the KL divergence losses for the audio and visual modalities can be defined as:
\begin{equation}
\mathcal{L}_\text{KL}^a = \sum_i p_s(y_i|\mathbf{x}_i) \log \frac{p_s(y_i|\mathbf{x}_i)}{p_a(y_i|\mathbf{x}_i)}
\end{equation}

\begin{equation}
\mathcal{L}_\text{KL}^v = \sum_i p_s(y_i|\mathbf{x}_i) \log \frac{p_s(y_i|\mathbf{x}_i)}{p_v(y_i|\mathbf{x}_i)}
\end{equation}

\subsubsection{Multi-Task Learning Optimization}

The HAEMSA framework incorporates a multi-task learning approach to leverage complementary information from related sentiment analysis tasks. For each task $t \in T$, a task-specific loss $L_t$ is defined based on the ground-truth labels and the predicted outputs.

\subsubsection{Joint Loss Function}

The total loss for training the modality-specific experts is a weighted combination of the KL divergence losses and the task-specific losses:
\begin{equation}
\mathcal{L} = \lambda_1 \mathcal{L}{\text{task}1} + \lambda_2 \mathcal{L}{\text{task}2} + \lambda_3 \mathcal{L}{\text{task}3} + \lambda_4 \mathcal{L}{\text{task}4} +\gamma(\mathcal{L}{\text{KL}}^{t} + \mathcal{L}{\text{KL}}^{a} + \mathcal{L}_{\text{KL}}^{v})
\end{equation}
where $\mathcal{L}_{\text{task}1}$, $\mathcal{L}{\text{task}2}$, $\mathcal{L}{\text{task}3}$, and $\mathcal{L}{\text{task}_4}$ are the task-specific losses, respectively, and $\lambda_1$, $\lambda_2$, $\lambda_3$, $\lambda_4$, and $\gamma$ are hyperparameters that control the balance between the task-specific losses and the KL divergence losses.

This joint optimization approach allows the HAEMSA framework to effectively learn from multiple modalities and tasks while ensuring that knowledge is shared across the hierarchical structure of experts.

\begin{table*}[h]
\centering
\small
\caption{Experimental results on IEMOCAP. The best results are highlighted in bold.}
\label{tab:iemocap_results}
\begin{tabular}{lccccccc}
\hline
Models & Happiness & Sadness & Neutral & Anger & Excitement & Frustration & Weighted-F1 \\
\hline
BC-LSTM (2017) & 34.43 & 60.87 & 51.81 & 56.73 & 57.95 & 58.92 & 54.95 \\
DialogueRNN (2019)& 33.18 & 78.80 & 59.21 & 65.28 & 71.86 & 58.91 & 62.75 \\
DialogueGCN (2019)& 51.87 & 76.76 & 56.76 & 62.26 & 72.71 & 58.04 & 63.16 \\
IterativeERC (2020)& 53.17 & 77.19 & 61.31 & 61.45 & 69.23 & 60.92 & 64.37 \\
QMNN (2021)& 39.71 & 68.30 & 55.29 & 62.58 & 66.71 & 62.19 & 59.88 \\
MMGCN (2021)& 42.34 & 78.67 & 61.73 & 69.00 & 74.33 & 62.32 & 66.22 \\
MVN (2022)& 55.75 & 73.30 & 61.88 & 65.96 & 69.50 & 64.21 & 65.44 \\
UniMSE (2022)& - & - & - & - & - & - & 70.66 \\
MultiEMO (2023)& 65.77 & 85.49 & 67.08 & 69.88 & 77.31 & 70.98 & 72.84 \\
UniMEEC (2024)& 69.52 & 88.51 & 69.74 & 72.63 & 78.80 & 72.98 & 74.83 \\
\hline
HAEMSA (Ours)& \textbf{71.95} & \textbf{91.61} & \textbf{72.18} & \textbf{75.17} & \textbf{81.56} & \textbf{75.53} & \textbf{77.67} \\
\hline
\end{tabular}
\end{table*}


\section{Experiments}
To evaluate the effectiveness of the proposed HAEMSA framework, we conduct extensive experiments on benchmark datasets for multimodal sentiment analysis. In this section, we describe the datasets, experimental setup, and results.

\subsection{Datasets}
We utilize three widely adopted datasets for multimodal sentiment analysis. CMU-MOSI \cite{zadeh_multi-attention_2018} contains 2,199 video clips from 93 speakers, annotated for sentiment on a scale from -3 to +3, split into 1,284 training, 229 validation, and 686 testing clips. CMU-MOSEI \cite{zadeh_multi-attention_2018}, an extension of CMU-MOSI, comprises 23,454 clips from 1,000 speakers, with 16,326 for training, 1,871 for validation, and 4,659 for testing. IEMOCAP \cite{busso2008iemocap} includes 12 hours of audiovisual data from 10 actors, featuring 10,039 utterances annotated with categorical emotions and dimensional attributes, divided into 5 sessions. All datasets provide pre-extracted features for text, audio, and visual modalities.






 \subsection{Experimental Setup}
We compare the performance of the HAEMSA framework with several state-of-the-art methods for multimodal sentiment analysis, including:

\begin{itemize}
\item  TFN \cite{zadeh2018multimodal}: Uses tensor factorization for multimodal fusion.
\item   LMF \cite{liu2018efficient}: Employs low-rank tensors for efficient multimodal fusion.
\item   MulT \cite{tsai2019multimodal}: Leverages cross-modal transformers for modality translation.
\item Self-MM \cite{yu2021learning}: The Self-Supervised Multi-task Multimodal sentiment analysis network incorporates an unimodal label generation module, grounded in self-supervised learning principles, to investigate the potential of unimodal supervision.
\item  MMIM \cite{han2021improving}: 
The MultiModal InfoMax framework introduces a stratified approach for maximizing mutual information, facilitating the model's capacity to cultivate cohesive representations across various modalities.
\item   TFR-Net \cite{yuan2021transformer}: The Transformer-based Feature Reconstruction Network integrates intra- and inter-modal attention mechanisms alongside a feature reconstruction component to effectively address the issue of sporadic missing features in non-aligned multimodal sequences.
\item  AMML \cite{sun2022learning}: The Adaptive Multimodal Meta-Learning framework employs a methodology derived from meta-learning to refine unimodal representations, which are subsequently adjusted to enhance multimodal integration.

\item EMT \cite{sun2023efficient}: Utilizes dual-level feature restoration techniques to enhance the robustness and accuracy of multimodal sentiment analysis.
\item CRNet \cite{shi2024co}: Interacts coordinated audio-visual representations to construct enhanced linguistic representations for improved sentiment analysis.

\item  BC-LSTM \cite{Poria2017}: A Bidirectional Contextual LSTM model that captures context in conversations to improve emotion recognition.
    \item DialogueRNN \cite{Majumder2019}: An attentive RNN designed for tracking emotional dynamics across conversational turns.
    \item DialogueGCN \cite{Ghosal2019}: Employs a Graph Convolutional Network to model dialogue structure, enhancing emotion detection.
    \item IterativeERC \cite{Lu2020}: An iterative network that refines emotion recognition predictions through successive stages.
    \item QMNN \cite{Li2021}: Utilizes quantum computing concepts to enhance conversational emotion recognition by capturing complex relationships in data.
    \item MMGCN \cite{Hu2021}: A Multimodal Graph Convolution Network that integrates various modalities to improve emotion recognition accuracy in conversations.
    \item MVN \cite{Ma2022}: Combines different perspectives for real-time emotion recognition, leveraging multiple data views for enhanced analysis.
    \item UniMSE \cite{Hu2022}: Aims for unified multimodal sentiment analysis and emotion recognition, focusing on consistent performance across various metrics.
    \item MultiEMO \cite{shi-huang-2023-multiemo}: A novel attention-based relevance-aware multimodal fusion framework that effectively integrates cross-modal cues. 
    \item UniMEEC \cite{hu2024unimeec}: A framework innovatively unifies multimodal emotion recognition in conversation (MERC) and multimodal emotion-cause pair extraction (MECPE) by reformulating them as mask prediction tasks, leveraging shared prompt learning across modalities, and employing task-specific hierarchical context aggregation.

\end{itemize}




To ensure robustness, we report results averaged over ten runs with different random seeds. This rigorous experimental setup allows for a fair and comprehensive evaluation of HAEMSA against a wide spectrum of existing approaches in the field. For MOSI and MOSEI, we report the binary accuracy (Acc-2), 7-class accuracy (Acc-7) and F1 score. For IEMOCAP, we report the weighted-F1 and unweighted accuracy (UA) for emotion classification. Weighted-F1 is the weighted average of F1 scores for all emotion categories while UA is the average accuracy of each class.


\begin{table*}[h]
\centering
\caption{Ablation study results on CMU-MOSEI and CMU-MOSI datasets}
\label{tab:ablation_study}

\begin{tabular}{lcccccccc}
\hline
\multirow{2}{*}{Method} & \multicolumn{4}{c}{CMU-MOSEI} & \multicolumn{4}{c}{CMU-MOSI} \\
& Acc-7(\%) & Acc-5(\%) & Acc-2(\%) & MAE & Acc-7(\%) & Acc-5(\%) & Acc-2(\%) & MAE \\
\hline
HAEMSA (Full) & \textbf{57.1} & \textbf{58.2} & \textbf{86.7} & \textbf{0.482} & \textbf{53.7} & \textbf{55.2} & \textbf{86.9} & \textbf{0.654} \\
\quad w/o Hierarchy & 55.8 & 57.2 & 83.1 & 0.611 & 51.3 & 48.7 & 82.2 & 0.717 \\
\quad w/o Evolution & 53.6 & 56.1 & 82.8 & 0.682 & 52.1 & 46.6 & 82.0 & 0.773 \\
\quad w/o Cross-Modal Integration & 52.2 & 55.7 & 82.2 & 0.691 & 47.6 & 49.1 & 80.4 & 0.861 \\
\quad w/o MTL & 51.7 & 55.1 & 82.9 & 0.672 & 47.1 & 48.7 & 80.6 & 0.848 \\
\hline
\end{tabular}%

\end{table*}

\section{Results}
We present a comprehensive evaluation of the proposed HAEMSA framework, comparing its performance against state-of-the-art methods and analyzing the contribution of each component through ablation studies.

\subsection{ Comparison to State-of-the-art}

\begin{table*}[h]
\centering
\caption{Ablation study results on IEMOCAP dataset}
\label{tab:iemocap_ablation}

\begin{tabular}{lccccccc}
\hline
Method & Happiness & Sadness & Neutral & Anger & Excitement & Frustration & Weighted-F1 \\
\hline
HAEMSA (Full) & \textbf{71.95} & \textbf{91.61} & \textbf{72.18} & \textbf{75.17} & \textbf{81.56} & \textbf{75.53} & \textbf{77.67} \\
\quad w/o Hierarchy & 67.42 & 87.03 & 68.57 & 71.41 & 77.48 & 71.75 & 73.79 \\
\quad w/o Evolution & 65.26 & 85.17 & 66.79 & 69.65 & 75.62 & 69.99 & 71.93 \\
\quad w/o Cross-Modal Integration & 63.81 & 83.92 & 65.34 & 68.20 & 74.07 & 68.53 & 70.38 \\
\quad w/o MTL & 63.18 & 83.29 & 64.71 & 67.57 & 73.44 & 67.90 & 69.75 \\
\hline
\end{tabular}%

\end{table*}

As shown in Table \ref{tab:overall_performance} and \ref{tab:iemocap_results}, HAEMSA consistently outperforms existing methods across all datasets and metrics. On CMU-MOSEI, HAEMSA achieves a 2.6\% increase in Acc-7 and a 0.059 decrease in MAE compared to the next best method. For CMU-MOSI, we observe a 6.3\% improvement in Acc-7 and a 0.058 reduction in MAE. On IEMOCAP, HAEMSA surpasses the state-of-the-art by 2.84\% in weighted-F1 score. These results demonstrate HAEMSA's effectiveness in capturing complex multimodal interactions and generalizing across different emotional contexts.




\subsection{Ablation Study}
To gain deeper insights into the efficacy of each component within the HAEMSA framework, we conducted comprehensive ablation studies across three benchmark datasets: CMU-MOSEI, CMU-MOSI, and IEMOCAP. This systematic analysis aims to quantify the individual contributions of key components and their synergistic effects on the overall performance of our proposed model. The detailed results of these ablation experiments are presented in Tables \ref{tab:ablation_study} and \ref{tab:iemocap_ablation}, offering a nuanced view of how each element impacts the model's effectiveness in multimodal sentiment analysis and emotion recognition tasks.

\begin{figure*}[ht!]
\centering
\includegraphics[width=0.75\textwidth]{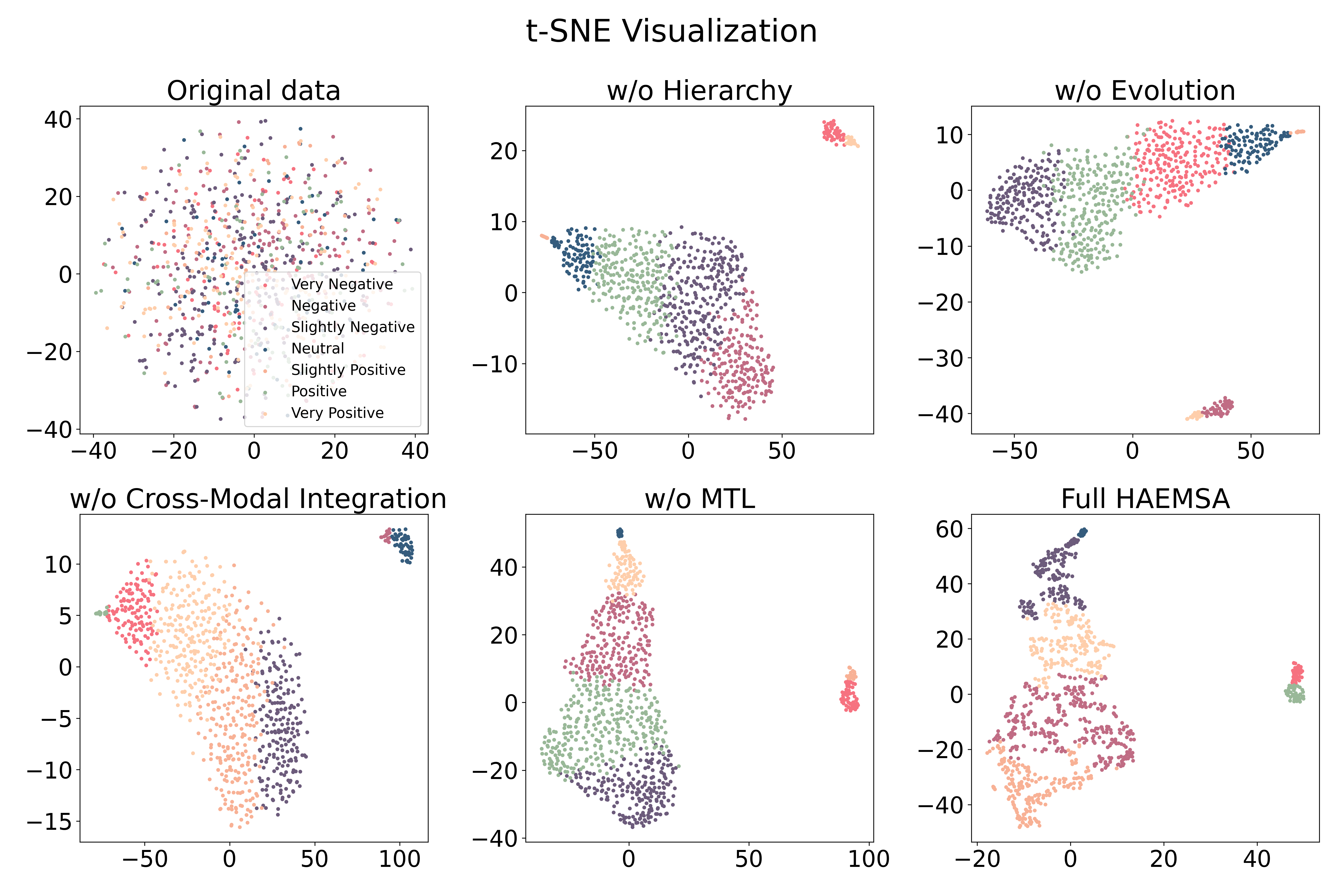}
\caption{t-SNE visualization of multimodal embeddings on CMU-MOSEI.}
\label{fig:tsne_visualization}
\end{figure*}

The ablation study conducted across CMU-MOSEI, CMU-MOSI, and IEMOCAP datasets consistently demonstrates the crucial role of each HAEMSA component in its overall performance. Results are presented in Tables \ref{tab:ablation_study} and \ref{tab:iemocap_ablation}. Removing the hierarchical structure (w/o Hierarchy) leads to substantial performance drops across all metrics and emotion categories, underscoring the importance of learning representations at different granularity levels. The evolutionary optimization process (w/o Evolution) proves vital, with its removal resulting in decreased performance, particularly evident in complex emotions like excitement and frustration on IEMOCAP. Both cross-modal integration and multi-task learning components show significant contributions, as their individual removals lead to notable performance degradation across all datasets. These findings highlight the synergistic effect of HAEMSA's components in effectively capturing and utilizing multimodal information for sentiment analysis and emotion recognition tasks.

\subsection{Visualization Analysis}
To provide insights into HAEMSA's learned representations, we visualize the multimodal embeddings using t-SNE. Figure \ref{fig:tsne_visualization} presents the t-SNE plots for different ablation settings.

The full HAEMSA model (Figure \ref{fig:tsne_visualization}(a)) shows clear separation between sentiment classes on CMU-MOSEI, indicating its ability to learn discriminative features. In contrast, ablated versions exhibit less distinct clusters, highlighting the importance of each component in learning effective representations.

\section{Discussion}
The HAEMSA framework, while demonstrating superior performance across multiple datasets, presents certain limitations that warrant discussion. The evolutionary optimization process, though effective, increases the computational cost compared to simpler fusion methods. This heightened complexity may restrict HAEMSA's applicability in resource-constrained or real-time scenarios, potentially limiting its use in certain practical applications.

Another consideration is the current implementation's assumption of complete modality availability for each data sample. In real-world applications, missing or incomplete modalities are common occurrences, presenting a challenge for the model's robustness and generalizability. This limitation highlights the need for adaptability in handling varied input conditions.

Despite significant improvements in emotion recognition, HAEMSA still faces challenges in distinguishing between closely related emotional states, particularly evident in the IEMOCAP dataset results. This limitation points to the need for more nuanced emotional modeling and representation learning.

Furthermore, the model's performance in cross-lingual or cross-cultural settings remains unexplored. Given the diverse nature of emotional expression across cultures, this aspect is crucial for the broader applicability of HAEMSA in global contexts. These limitations, while not diminishing the model's achievements, provide clear directions for future research and improvement.

\section{Conclusion}

This paper introduces HAEMSA (Hierarchical Adaptive Expert for Multimodal Sentiment Analysis), a novel framework that integrates hierarchical mixture-of-experts networks, evolutionary optimization, and multi-task learning for multimodal sentiment analysis and emotion recognition. HAEMSA's hierarchical structure enables learning at various levels of modality granularity, while its evolutionary process optimizes expert architectures for effective cross-modal knowledge transfer. The incorporation of multi-task learning allows HAEMSA to capture task interdependencies, potentially improving overall performance.

Experiments on CMU-MOSI, CMU-MOSEI, and IEMOCAP datasets demonstrate HAEMSA's effectiveness compared to existing methods, showing improvements in accuracy and MAE across different tasks. These results suggest that HAEMSA can effectively capture complex multimodal interactions and generalize across various emotional contexts.

Future work could focus on addressing current limitations and expanding HAEMSA's capabilities. This may include optimizing computational efficiency to enhance real-world applicability, developing mechanisms for handling missing or incomplete modalities, and refining the model's ability to distinguish between nuanced emotional states. Additionally, exploring HAEMSA's performance in cross-lingual and cross-cultural contexts could broaden its impact. Such advancements could contribute to the wider field of multimodal sentiment analysis and emotion recognition, potentially opening new avenues for applications in areas such as human-computer interaction, mental health monitoring, and social media analysis.

{\small
\bibliographystyle{ieee_fullname}
\bibliography{wacv2025cikm}
}

\end{document}